\documentclass{article}

\usepackage{microtype}
\usepackage{graphicx}
\usepackage{subfigure}
\usepackage{amsfonts}
\usepackage{amsmath}
\usepackage{booktabs}

\usepackage[ruled,vlined]{algorithm2e}

\usepackage[font=small,skip=0pt]{caption}
\usepackage{hyperref}

\usepackage[accepted]{icml2020}

\icmltitlerunning{Model-based Meta Reinforcement Learning using Graph Structured Surrogate Models}

\begin{document}

\twocolumn[
\icmltitle{Model-based Meta Reinforcement Learning using\\ Graph Structured Surrogate Models}

% It is OKAY to include author information, even for blind
% submissions: the style file will automatically remove it for you
% unless you've provided the [accepted] option to the icml2020
% package.

% List of affiliations: The first argument should be a (short)
% identifier you will use later to specify author affiliations
% Academic affiliations should list Department, University, City, Region, Country
% Industry affiliations should list Company, City, Region, Country

% You can specify symbols, otherwise they are numbered in order.
% Ideally, you should not use this facility. Affiliations will be numbered
% in order of appearance and this is the preferred way.
\icmlsetsymbol{equal}{*}

\begin{icmlauthorlist}
\icmlauthor{Qi Wang}{to}
\icmlauthor{Herke van Hoof}{to}
\end{icmlauthorlist}

\icmlaffiliation{to}{Amsterdam Machine Learning Lab, University of Amsterdam, Amsterdam, the Netherlands}

\icmlcorrespondingauthor{Qi Wang}{hhq123go@gmail.com}

% You may provide any keywords that you
% find helpful for describing your paper; these are used to populate
% the "keywords" metadata in the PDF but will not be shown in the document
\icmlkeywords{Machine Learning, ICML}

\vskip 0.3in]

% this must go after the closing bracket ] following \twocolumn[ ...

% This command actually creates the footnote in the first column
% listing the affiliations and the copyright notice.
% The command takes one argument, which is text to display at the start of the footnote.
% The \icmlEqualContribution command is standard text for equal contribution.
% Remove it (just {}) if you do not need this facility.

%\printAffiliationsAndNotice{}  % leave blank if no need to mention equal contribution
\printAffiliationsAndNotice{} % otherwise use the standard text.

\begin{abstract}
Reinforcement learning is a promising paradigm for solving sequential decision-making problems, but \textit{low data efficiency} and \textit{weak generalization} across tasks are bottlenecks in real-world applications. 
Model-based meta reinforcement learning addresses these issues by learning dynamics and leveraging knowledge from prior experience. 
In this paper, we take a closer look at this framework, and propose a new Thompson-sampling based approach that consists of a new model to \textit{identify task dynamics} together with an \textit{amortized policy optimization} step.  
We show that our model, called a graph structured surrogate model (GSSM), outperforms state-of-the-art methods in predicting environment dynamics. Additionally, our approach is able to obtain high returns, while allowing \textit{fast execution} during deployment by avoiding test-time policy gradient optimization. 
\end{abstract}

%introduce the significance of MBMRL, current research progress and bottlenecks, main novelty in our model and a specified application scenario.
\section{Introduction}
Reinforcement learning (RL) has been successfully applied to several complicated tasks and achieved remarkable performance, even surpassing outstanding human players in a variety of domains
\cite{mnih2015human,silver2017mastering,vinyals2019alphastar}. By exploration and exploitation, a series of sequential decision-making problems can be theoretically addressed in this paradigm. 

As a cutting-edge research topic, there still remain long standing challenges when putting RL into practice. In principle, these can be overviewed from three aspects: i) \textit{data efficiency}, the prevalent branch of RL algorithms as model free reinforcement learning (MFRL) poses great demands on massive interactions with an environment, making it unrealistic to conduct in most real-world applications \cite{sutton2018reinforcement,chua2018deep}. ii) \textit{robustness to unseen environments}, when an environment of interest drifts in terms of dynamics or reward mechanisms, previously learned skills suffer the risk of poor generalization \cite{jing2018learning,clavera2019learning}. And dynamics mismatch easily leads to \textit{Sim2Real} problems \cite{peng2018sim}. iii) \textit{instantaneously planning}, either policy learning or calibration in policies consumes additional time in execution phases \cite{wang2019exploring}. And critical issues might arise in real-time planning missions with strict constraints like autonomous driving.

To address above mentioned fundamental concerns, we propose a graph structured surrogate model (GSSM) within the framework of Model-based Meta Reinforcement Learning (MBMRL) \cite{nagabandi2019learning,saemundsson2018meta,killian2017robust,lee2020context}. 
Our approach to optimize GSSM is built on the principle of posterior sampling \cite{osband2013more}, and more efforts are made to improve \textit{dynamics forecasting}, accelerate \textit{policy learning} and enable \textit{fast adaptations} across tasks via latent variables. Especially, unlike most existing MBMRL algorithms using time-expensive \textit{derivative-free} algorithms in model predictive control, we explore to learn \textit{amortized policies} for corresponding tasks, which do not require adaptation time when faced with a new task in a policy sense. 

As a preliminary trial in this domain, associating dynamics models and policies with task-specific latent variables paves a promising path to attain presumed goals in MBMRL.
Our primary contributions are summarized as follows:
\begin{itemize}
    \item A graph structured dynamics model is developed with superior generalization capability across tasks, which enables effectively \textit{encoding of memories} and \textit{abstracting environments} in a latent space.
    
    \item We explore a new strategy for meta model-based policy search that allows for latent variables' participation in policy networks to achieve \textit{fast adaptations} to new tasks without \textit{additional policy gradient updates}.
    
    \item Extensive experiments on a variety of tasks demonstrate superior performance of GSSM in terms of both generalization capability and sample complexity by comparing with other typical algorithms.
\end{itemize}

%literature works summary: model based rl for data efficiency, meta learning for skill transfer, combination for uncertainty-aware and data efficient rl
\section{Literature Review}
As already mentioned, several critical bottlenecks restrict universal applications of RL algorithms. 
In terms of mastering new skills rapidly, meta learning is an ideal paradigm to achieve with a few instances.
As for data efficiency, both model-based reinforcement learning (MBRL) and meta learning can reduce sample complexity. 

\textbf{Meta Learning.} The core of meta learning is a paradigm to discover implicit common structures across a collection of similar tasks and then generalize such knowledge to new scenarios. 
%In other words, it learns to learn the strategies of leveraging previous or prior knowledge to transfer skills.
%One feasible schema is to disentangle the meta (global) parameters and task-specific (local) parameters in jointly learning a batch of tasks each time. 
Leveraging knowledge from a meta learner to a task-specific learner is called \textit{fast adaptation}. 
Two strategies are quite popular for meta learning, respectively as Gradient-based Meta Learning and Contextual Meta Learning. 
A representative framework for gradient-based meta learning is model agnostic meta learning (MAML) \cite{finn2018probabilistic,finn2017model,yoon2018bayesian,flennerhag2019meta,lee2018gradient}, where both a meta learner and an adaptor are derived via gradient information after a few shots in a specific task. 
The contextual meta learning algorithms rely on task specific latent variables to identify a task after a few observations. 
This strategy theoretically does not require gradient adaptations in new tasks but constructing task relevant latent variables are decisive \cite{garnelo2018conditional,garnelo2018neural,hausman2018learning,kim2019attentive}.

\textbf{Model-based Reinforcement Learning.} Key to applications within a RL framework is how to boost sample efficiency, and MBRL serves a role as approximating a target environment as close as possible. 
In an environment with unknown dynamics, MBRL mainly investigates a deterministic map or a distribution $p:\mathcal{S}\times\mathcal{A}\to\mathcal{S}$.
Generally, deterministic modelling on dynamical systems does not involve random variables in the hidden units, and some auto-regressive neural network structures are quite typical in this family \cite{leibfried2016deep,gehring2017convolutional,nagabandi2017neural,amos2018learning,van2020plannable}.
Stochastic modelling on dynamical systems are mainly formulated by incorporating uncertainty in system parameters and observation noise \cite{deisenroth2011pilco,kamthe2017data,eslami2018neural,hafner2018learning,chua2018deep,ha2018world}. 

\textbf{Meta Reinforcement Learning.} Most of meta RL algorithms follow a model-free paradigm, e.g. MAESN \cite{gupta2018meta}, CAVIA \cite{zintgraf2019fast} and PEARL \cite{rakelly2019efficient}.
The idea of marrying meta Learning and MBRL can further reduce sample complexity but is not widely studied. 
The work \cite{nagabandi2019learning} takes a gradient-based strategy as MAML and alleviates the gap of \textit{Sim2Real}. 
%Gaussian processes can also be used in Meta Learning tasks, since the predictive distribution with observations can carry out task-aware predictions and measure the uncertainty. 
In \cite{saemundsson2018meta}, Gaussian process latent variables work as task inference to learn dynamics in tasks. In particular, a model strongly related to ours is in \cite{galashov2019meta}, where neural processes (NPs) are used to identify dynamics of tasks, but it requires to re-train or fine-tune parameterized policies via gradient updates in new tasks.
Note that most of these MBMRL algorithms focus on \textit{fast adaptations} in \textit{dynamics models}. All of these either make use of \textit{derivative-free} algorithms for model predictive control or re-train policies in separate tasks, which are prohibitively \textit{time-consuming} in real-world planning problems \cite{wang2019exploring}.

\begin{figure*}[ht]
\vskip 0.1in
\begin{center}
\centerline{\includegraphics[width=0.90\textwidth]{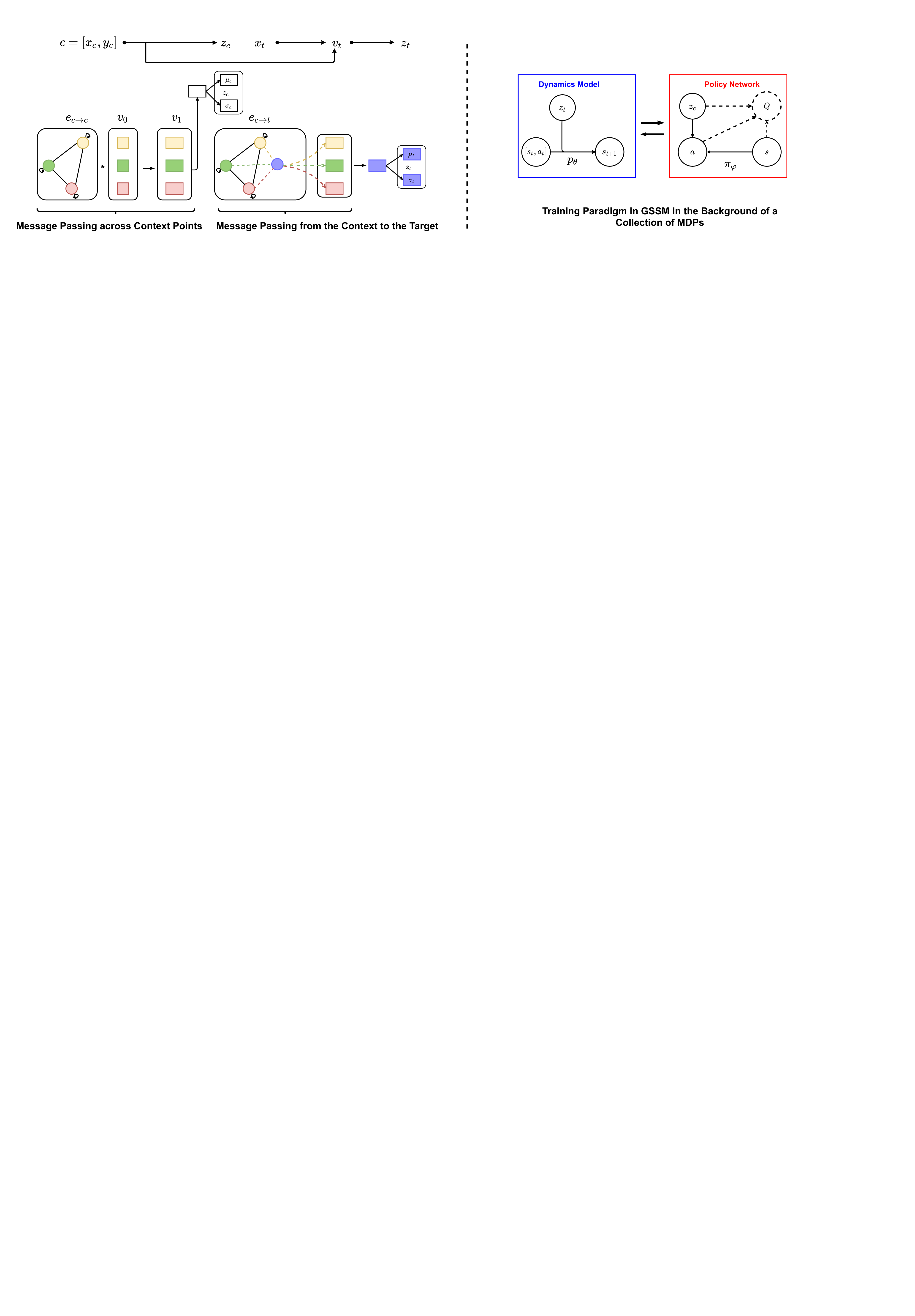}}
\caption{Graph Structured Surrogate Model. In the \textbf{Left} : Information Flows in \textcolor{blue}{\textbf{Dynamics Models}} from the left to the right describe the Message Passing between context points (Edges in solid lines) and to the \textcolor{blue}{\textbf{target point}} (Edges in dashed lines). In the \textbf{Right} : Latent variables participate in both \textcolor{blue}{\textbf{Dynamics Models}} and \textcolor{red}{\textbf{Policy Networks}} (Dashed elements are involved in the module when using Actor-Critic frameworks and Double arrows mean interactions to learn \textit{amortized policies}).}
\label{GSSM_Structure}
\end{center}
\vskip -0.3in
\end{figure*}

%formulate the research problems and supplement some preliminaries on meta reinforcement learning and predictive distribution using latent variables.
\section{Problem Formulation and Preliminaries}
The phase of decision-making in RL is usually characterized with a discrete-time Markov Decision Process (MDP), denoted as $\mathcal{M}$.
Given states $s_t\in\mathcal{S}$, actions $a_t\in\mathcal{A}$, policy functions $\mathcal{\pi}$, state transition distributions $\mathcal{P}$, reward feedback equations $\mathcal{R}$ and a discount factor $\mathcal{\gamma}$ for a step-wise reward, a MDP can be formalized with a tuple of these elements $\mathcal{M}_{k}=(\mathcal{S},\mathcal{A},\mathcal{\pi},\mathcal{P}_{k},\mathcal{R}_{k},\mathcal{\gamma})$. The return of accumulated rewards is a discounted summation of reward feedback $r(s_{t},a_{t})$ along the trajectories $\tau:=(s_0,a_0,r_0,\dots,s_{H-1},a_{H-1},r_{H-1},s_{H})$. 

Hence, maximizing the expected cumulative rewards over trajectories is the objective of policy optimization in classical RL problems.
In contrast, MBMRL considers a distribution over MDPs $\mathcal{M}\sim p(M)$, and the principal goal is to simultaneously \textit{construct and enable fast adaptations in dynamics models} and \textit{plan with learned dynamics models}. To this end, the paradigm is designed for diverse configurations of environments.

\subsection{Optimization Objective in MBMRL}
More formally, we can reformulate MBMRL problems from the insight of optimization and two correlated objectives are attached as follows.

\begin{equation}
    \begin{split}
        \max_{\theta}\mathbb{E}_{\mathcal{M}\sim p(M)\atop([s,a],s^{\prime})\sim \mathcal{M}}\ln\left[p_{\theta_{\mathcal{M}}}(s^{\prime}\vert [s,a])\right]\: \text{s.t.} \, p_{\theta_{\mathcal{M}}}=u(\theta,\mathcal{D}^{\text{tr}}_{\mathcal{M}})\\
        \max_{\varphi_{\mathcal{M}}}\mathbb{E}_{s^{\prime}\sim p_{\theta_\mathcal{M}}(s^\prime\vert[s,a])\atop a\sim\pi_{\varphi_{\mathcal{M}}}}\left[\sum_{t=0}^{H-1}\gamma^{t}r_{\mathcal{M}(s_t,a_t)}\right]
        \:\forall\mathcal{M}\sim p(M)
    \end{split}
    \label{meta_obj}
\end{equation}

Here the upper side term of Eq. (\ref{meta_obj}) is to maximize the log-likelihood of state-transitions $p(s^{\prime}\vert [s,a])$ in a collection of MDPs and $u$ means the \textit{fast adaptation} mechanism in $\mathcal{M}$ to learn an updated dynamics model $p_{\theta_{\mathcal{M}}}$ with meta-learned parameters $\theta$ and a few transition instances $\mathcal{D}^{\text{tr}}_{\mathcal{M}}$.
The lower side one corresponds to learning a policy or finding a planning strategy $\pi_{\varphi_{\mathcal{M}}}$ in separate dynamics models. 

It is worth noting that the objective of MBMRL differs from that in traditional MFMRL, since it consists of two phases as \textit{dynamics model learning} and \textit{policy optimization}. 

For more insights, we define the discrepancy between a distribution over MDPs and a distribution over learned dynamics models using the expected form of the total variance distance as $\mathbb{E}_{M\sim p(M)\atop (s,a)\sim\nu(s,a)}\big[D_{\text{TV}}[P_{\hat{M}}(\cdot\vert s,a)),P_{M}(\cdot\vert s,a)]\big]$, where $P_{\hat{M}}(\cdot\vert s,a)$ and $P_{M}(\cdot\vert s,a)$ are respectively a learned state transition distribution and a transition distribution in a real environment and $\nu(s,a)$ is a distribution over state action pairs. By extending findings in \cite{rajeswaran2020game} to meta learning, we can depict a performance gap between dynamics models and real environments under an arbitrary policy in \textbf{Lemma 1}.

\textbf{Lemma 1.} \textit{Assume the discrepancy between transition distributions over MDPs and the learned approximated models} $\mathbb{E}_{M\sim p(M)\atop (s,a)\sim\nu(s,a)}\big[D_{\text{TV}}[\hat{P}_{M}(\cdot\vert s,a),P_{M}(\cdot\vert s,a)]\big]\leq\epsilon$, \textit{we can estimate the performance gap under a policy} $\pi$ \textit{as follows},

\begin{equation}
    \begin{split}
        \mathbb{E}_{M\sim p(M)}\left[|\mathcal{J}_{\hat{M}}(\pi)-\mathcal{J}_M(\pi)|\right]\leq\frac{2\epsilon\mathcal{R}_{\max}}{(1-\gamma)^2}
    \end{split}
\end{equation}

\textit{where} $\mathcal{R}_{\max}$ \textit{is the supremum value of one step reward with discount factor} $\gamma$, $\hat{M}$ \textit{and} $M$ \textit{are a sampled approximated MDP and a corresponding real MDP, and the expected rewards are} $\mathcal{J}_{M}(\pi)=\mathbb{E}_{s^{\prime}\sim p_{\mathcal{M}}(s^\prime\vert[s,a])\atop a\sim\pi(\cdot\vert s)}\big[\sum_{t=0}^{\infty}\gamma^{t}r_{\mathcal{M}(s_t,a_t)}\big]$.

\textbf{Theorem 1.} \textit{Suppose a bounded dynamics models' discrepancy as} $\mathbb{E}_{M\sim p(M)\atop (s,a)\sim\nu(s,a)}\big[D_{\text{TV}}[\hat{P}_{M}(\cdot\vert s,a),P_{M}(\cdot\vert s,a)]\big]\leq\epsilon$ \textit{over a MDP distribution and optimal policies for any MDP} $\mathcal{M}$ \textit{and its approximation} $\mathcal{\hat{M}}$ \textit{are respectively} $\pi_{\hat{M}}$ \textit{and} $\pi_M$. \textit{Then the regret bound is estimated as follows}.

\begin{equation}
    \begin{split}
        \mathbb{E}_{M\sim p(M)}\left[\mathcal{J}_{M}(\pi_{\hat{M}})\right]\geq
        \mathbb{E}_{M\sim p(M)}\left[\mathcal{J}_{M}(\pi_{M})\right]
        -\frac{4\epsilon\mathcal{R}_{\max}}{(1-\gamma)^2}
    \end{split}
    \label{mbmrl_bound}
\end{equation}

The proposition in \textbf{Lemma 1} is to disentangle relationship between dynamics learning and policy performance with a performance difference, suitable for most MBMRL algorithms. In addition, the inherent model bias in MBRL tends to bring catastrophic failure, and in meta learning scenarios \textbf{Theorem 1} implies that potential expected regret can be well bounded or minimized if the discrepancy in a dynamics model sense is small enough. The proof is given in Appendix (\ref{Append_A}/\ref{Append_B}).

\subsection{Meta Learning with Latent Variables}
In the framework of meta learning, the core is to achieve a rapid skill transfer from previous domains to new domains. 
Distinguished from gradient-based methodologies, task-specific latent variable models are appealing recently. 
A latent variable $z$, which summarizes statistics of a specific environment $\mathcal{M}$, can be inferred from a few shots of instances \cite{garnelo2018neural,garnelo2018conditional}. 
The inferred latent variables can participate in either dynamics models $p_{\theta}(\Delta s\vert [s,a],z)$ \cite{galashov2019meta,lee2020context} or policy learning $\pi_{\varphi}(a\vert s,z)$ \cite{rakelly2019efficient} phases.
The former one matches \textit{system identification}, while the latter indicates \textit{task-specific policies}. No gradient steps are required in testing time due to the use of latent variables.
Apparently, we need to design a flexible non-parametric approximation for a probabilistic task embedding and transformations should preserve crucial informative traits. 

%introduce the meta dynamics model and policy learning in our paradigm
\section{Graph Structured Surrogate Models}
Similar to the work \cite{galashov2019meta}, we learn task-specific dynamics models via contextual latent variables. Importantly, these variables are also allowed to participate in policy networks to enable fast \textit{planning} and \textit{adaptations} without gradient updates as displayed in the left side of Fig (\ref{GSSM_Structure}). Throughout GSSM, a posterior sampling strategy \cite{osband2013more} is employed, which \textit{samples new contextual variables}, \textit{executes amortized policies} and \textit{updates posteriors of latent variables} in a loop. 
For the sake of simplicity, notations to describe transitions in dynamics models are simplified using $x=[s,a]$, $y=s^{\prime}$ or $y=\Delta s$ in following sections (notations used in Fig. (\ref{GSSM_Structure})).

\subsection{Graph Structured Latent Variables}
In comparison to Neural Processes \cite{garnelo2018neural}, which impose \textit{a simple pooling} operation over context points for latent variables, we propose to model the representation of latent variables via Graph Neural Networks \cite{kipf2016semi,satorras2018few,wang2018nervenet}.
The dependencies between the context and target transition samples are reasoned in a latent space and informative templates can be selected from memories \cite{bornschein2017variational}, improving predictive performance in dynamics.

For transition dataset from a task, we have a set of context points $[x_c,y_c]$ and the target point $[x_*,y_*]$. We treat the context points in a form of graph structured dataset $\mathcal{G}=<\mathcal{V},\mathcal{E}>$, which comprises of a collection of vertices $\mathcal{V}=[x_c,y_c]$ and relational edges $\mathcal{E}\subseteq\mathcal{V}\times\mathcal{V}$. Then the graph convolution operator over any data point $x_*$ can be defined,
\begin{equation}
\begin{split}
     f(\mathcal{V}_{\mathcal{C}})=\sigma\left(D^{-1}L[\mathcal{V}_{\mathcal{C}}\mathcal{W}]\right), \\
     \quad \mathcal{G}(x_*)\circ f=\sum_{i\in\mathcal{N}_*}f(v_{i})\texttt{sim}(x_*,x_{i})
    \label{gnn}   
\end{split}
\end{equation}
where $\mathcal{V}_{\mathcal{C}}$ is the feature matrix of the context points, the weight parameter is $\texttt{sim}(x_*,x_{i})$ and $f(v_{i})$ is the embedding of a node in the graph $\mathcal{G}$ after message passing processes from its neighbors.
The graph Laplacian matrix $L$ reveals the connection relationship in a normalized way \cite{kipf2016semi}, where $D$ is to normalize the row elements in Eq. (\ref{gnn}). 
The value as the initial input for each node is flexible, and available forms can be either $[x_i,y_i]$ or direct $y_i$.
Here a fully connected graph is built to characterize the pairwise relationship, and the construction of the graph Laplacian matrix is based on the pairwise similarities between instances $\texttt{sim}(x_i,x_j)=\frac{\langle t(x_i),t(x_j)\rangle}{\|t(x_i)\|_2\cdot\|t(x_j)\|_2}$, where $t(x)$ is the intermediate transformation of features for a sample $x$ and the notation $\langle\cdot,\cdot\rangle$ means a dot product. All of these are well illustrated in the right side of Fig. (\ref{GSSM_Structure}).

\textbf{Message Passing between Context Points.} This process is to deliver and aggregate neighborhood information for node representations, which can be specified via a constructed Laplacian matrix,
\begin{equation}
    \begin{split}
        \hat{s}_{ij}=\frac{\exp{(\beta\cdot \texttt{sim}(x_i,x_j))}}{\sum_{j\in\mathcal{N}_{i}}\exp{\left(\beta\cdot \texttt{sim}(x_i,x_j)\right)}},\\
        h_i^{(l+1)}=\sigma\left(W^{(l)}h_i^{(l)}+\sum_{j\in\mathcal{N}_{i}}\hat{s}_{ij}W^{(l)}h_j^{(l)}\right)
    \end{split}
    \label{c_embedding}
\end{equation}
where $\beta$ is a tunable parameter for pairwise similarities, $W^{(l)}$ is the network parameter for node feature transformations in $l$-th layer, $\hat{s}_{ij}$ is the normalized Laplacian weight, and $h_j^{(l)}$ is the $l$-th intermediate representation after message passing from sample $i$'s neighborhoods $\mathcal{N}_{i}$ to itself.
Eq. (\ref{c_embedding}) is an instantiation for Eq. (\ref{gnn}) with self-loop information propagation.
The node embeddings after the final message passing are denoted with $h$ in the following section. Due to the operation of message passing in Eq. (\ref{mp_intra}), a learned representation for each node summarizes statistics of interactions and could be more robust in some sense,

\begin{equation}
    \sum_{j\in\mathcal{N}_i}\hat{s}_{ij}h_{j}= r_{j},\bigoplus_{i\in\mathcal{V}}r_{i}=r_c,
    \quad q(z_c)=\mathcal{N}\big(\mu(r_c),\Sigma(r_c)\big)
    \label{mp_intra}
\end{equation}

where $\bigoplus$ means pooling operations over all node vectors and the last term describes the \textit{amortized} distribution \cite{zhang2018advances} for context points.

\textbf{Message Passing from the Context to the Target.} This process is to transmit task-beneficial information from the context to the target. And consider the contributions to the target vary from instance to instance, the aggregated context message $r_*$ can be represented in a weighted way, where the coefficient for each context point is computed in the same way as $\hat{s}(x_i,x_*)$ to measure the relevance.
\begin{equation}
    \begin{split}
        \sum_{i\in\mathcal{N}_*}\hat{s}(x_i,x_*)h_i=r_{*},\quad
        q_{\phi}(z_*)=\mathcal{N}\big(\mu(r_*),\Sigma(r_*)\big)
    \end{split}
    \label{t_embedding}
\end{equation}
After the message passing from the context to the target, $r_*$ is further mapped into mean and variance parameters of a proposal distribution $q_\phi(z_*)$ using neural networks, as displayed in the right side of Eq. (\ref{t_embedding}). Here we employ mean field amortized inference, using diagonal Gaussian distributions for the convenience of computations.  

\subsection{Approximate Inference and Scalable Training}
To learn dynamics with latent variables, we need to specify a predictive distribution and an objective in optimization. Here the data point $[x_c,y_c,x_*,y_*]$ coupled of the context $[x_c,y_c]$ and the target $[x_*,y_*]$ is sampled from meta learning dataset $p(D)$. Note that the predictive distribution $p(y_*\vert x_*,z_*)$ including a latent variable $z_*$ is unknown. 

Though the exact inference for the predictive distribution is intractable, one plausible way is to use the above mentioned variational distribution $q_{\phi}(z_*\vert x_*,x_c,y_c)$ in Eq. (\ref{t_embedding}).
As a result, the evidence lower bound (ELBO) is formulated in Eq. (\ref{dm_elbo}).

\begin{equation}
    \begin{split}
        \mathbb{E}_{p(D)}\big[\ln p(y_*\vert x_*,x_c,y_c)\big]\geq\mathbb{E}_{p(D)}\big[\mathbb{E}_{q_{\phi}}[\ln p_{\theta}(y_*\vert x_*,z_*)] \\
        -D_{KL}[q_{\phi}(z_*\vert x_*,x_c,y_c)\parallel p(z_*)]\big]
    \end{split}
    \label{dm_elbo}
\end{equation}

To well specify a dynamical system from context points, a variational distribution $q_{\phi_2}(z_c\vert x_c,y_c)$ is selected as a prior distribution $p(z_*)$ in ELBO , which is achieved by making Eq. (\ref{mp_intra}) a learnable multivariate diagonal Gaussian distribution. And the induced objective is as follows.

\begin{equation}
    \begin{split}
        \mathbb{E}_{p(D)}\big[\ln p(y_*\vert x_*,x_c,y_c)\big]\geq\mathbb{E}_{p(D)}\big[\mathbb{E}_{q_{\phi_1}}[\ln p_{\theta}(y_*\vert x_*,z_*)] \\
        -D_{KL}[q_{\phi_1}(z_*\vert x_*,x_c,y_c)\parallel q_{\phi_2}(z_c\vert x_c,y_c)]\big]
    \end{split}
    \label{dm_fina_elbo}
\end{equation}

When implementing ELBO in practice, Monte Carlo estimation is performed for the negative form of right side of Eq. (\ref{dm_fina_elbo}),
\begin{equation}
    \begin{split}
        \mathcal{L}(\theta)=-\frac{1}{K}\sum_{t=1}^{T}\sum_{k=1}^{K}\ln p_{\theta}(y^{(t)}_*\vert x^{(t)}_*,z_*^{(t,k)}) \\
        +D_{KL}\big[q_{\phi_1}(z^{(t)}_*\vert x^{(t)}_*,x^{(t)}_c,y^{(t)}_c)\parallel q_{\phi_2}(z^{(t)}_c\vert x^{(t)}_c,y^{(t)}_c)\big]
    \end{split}
    \label{dm_mc_elbo}
\end{equation}
where $T$ is the batch size of samples in meta training, $K$ is the number of particles in estimation, and latent variable values are sampled from the approximate posterior $z_*^{(t,k)}\sim q_{\phi_1}(z^{(t)}_*\vert x^{(t)}_*,x^{(t)}_c,y^{(t)}_c)$.

Similarly, when it comes to prediction using the learned dynamics model, the Monte Carlo estimator can be directly applied again to derive a predictive distribution in Eq. (\ref{md_predictive}) with the approximate posterior $q_{\phi_1}$ and collected context points $[x_c,y_c]$.
\begin{equation}
\begin{split}
    p(y_*\vert x_*,x_c,y_c)=\int q_{\phi_1}(z_*\vert x_*,x_c,y_c)p_{\theta}(y_*\vert x_*,z_*)dz_* \\
    \approx
    \frac{1}{K}\sum_{k=1}^{K}p_{\theta}(y_*^{(k)}\vert x_*,z_*^{(k)})
\end{split}
\label{md_predictive}
\end{equation}

\subsection{Amortized Policy Learning in Dynamical Systems}
Once a dynamics model is learned, planning can be achieved by interacting with the learned dynamics model.
Importantly, more focus is placed on \textit{fast adaptations of policies} instead of \textit{dynamics} across tasks here. The point is seldom explored in this domain.
We utilize \textit{posterior sampling} \cite{osband2013more} in capturing task-specific policies, which can be viewed as an approximation to Bayes-adaptive RL with good exploration properties. This is also known as Thompson sampling \cite{thompson1933likelihood} in Bandit cases. 

As a result, latent variables $z^{(t)}_c$ are sampled from the posterior belief $q_{\phi_2}(z^{(t)}_c\vert x^{(t)}_c,y^{(t)}_c)$ over tasks and rewards are maximized afterward. The maximization process, either re-training or fine-tuning meta-learned policies $\pi_\varphi(a\vert s)$ in dynamical systems, is computationally expensive in the work \cite{galashov2019meta}. So we \textit{amortize} this step by optimizing latent variable conditioned policies $\pi_\varphi(a\vert s,z^{(t)}_c)$ to predict task-specific maximization results.

Here we can parameterize a policy and get the dynamics model and the policy network connected in optimization with back-propagation through time (BPTT) \cite{kurutach2018model,parmas2018pipps}. In detail, this phase collects trajectories from a learned dynamics model and evaluates  rewards of policies in Eq. (\ref{meta_pg}) to maximize, and BPTT is extended to meta model-based policy search in this work. 

\begin{figure*}[ht]
\centering
\includegraphics[width=1.0\textwidth]{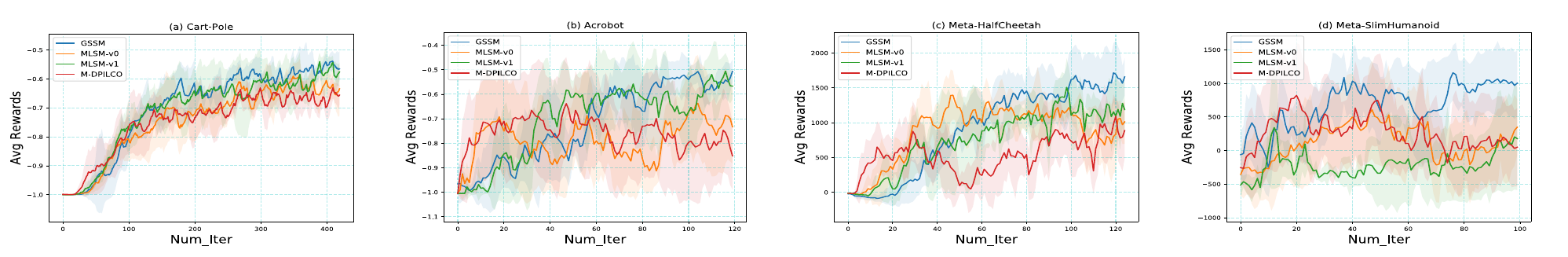}
\vspace*{-5mm}
\caption{Performance of Policies in Meta-Training Processes. Here environments are varied in terms of dynamics during iterations and each \texttt{iter} corresponds to one new sampled \texttt{episode} added in optimization. The average rewards are tested after each \texttt{iter} in an offline way and results indicate means and corresponding standard deviations in 5 runs.}
\label{Meta_train_results}
\end{figure*}

\begin{equation}
    \begin{split}
        \mathcal{J}(\varphi)=\int \mathcal{R}_{\tau}\big[p(s_0)\prod_{t=1}^{T}p_{\theta}(s_{t+1}\vert s_t,\pi_{\varphi},z_t)\big]dz_{1:T}ds_{0:T} \\
        \approx  \frac{1}{K}\sum_{k=1}^{K}\mathcal{R}_{\tau}^{(k)}=
        \frac{1}{K}\sum_{k=1}^{K}\sum_{t=0}^{T}\gamma^{t}r(s_{t+1}^{(k)})
    \end{split}
    \label{meta_pg}
\end{equation}

Apart from policy search strategies like BPTT, MFRL algorithms can be directly combined with the learned dynamics model as well. 
We extend this approach to \textit{actor-critic} frameworks, in which case the value function is also conditioned on the latent variable and a standard loss function for the value output is added to the objective.
For the sake of simplicity, we do not rewrite the objective for model-free algorithms in policy learning as Eq. (\ref{meta_pg}).  
Illustrations of this setup are given in the right side of Fig. (\ref{GSSM_Structure}).

%Meta training Algorithm in most MBMRLs in this paper
\begin{algorithm}[H]
\SetAlgoLined
\SetKwInOut{Input}{Input}
\SetKwInOut{Output}{Output}
\Input{MDP distribution $\rho(\mathcal{M})$ ;
Number of episodes $T$ ;
Exploration policy of dynamics $\pi_e$.}
\Output{Meta-trained parameters $\phi_*$, $\theta_*$ and $\varphi_*$.}
Initialize model parameters $\phi$, $\theta$ and $\varphi_*$ \;

\While{\text{Meta-Training not Completed}}{
Sample a task $\mathcal{M}_k$ from a distribution $\rho(\mathcal{M})$\;

Perform $T$ roll-outs to collect $\mathcal{D}_k$ with $\pi_e$\;

Optimize $\phi$ and $\theta$ on $\mathcal{D}_k$ in Eq. (\ref{dm_mc_elbo})\;

\For{$i=1,2,\dots,N$}{
Sample initial states from the learned $\text{DM}_*$\;

Collect episodes in $\text{DM}_*$ using $\pi_{\varphi}$\;

Evaluate cumulative rewards in Eq. (\ref{meta_pg})\;

Optimize the policy $\pi_{\varphi}$: $\varphi\leftarrow\varphi+\alpha\nabla_{\varphi}\mathcal{J}$.}
}
\caption{Meta-Training Phases.}
\label{Meta_Tr}
\end{algorithm}
%Meta training Algorithm in most MBMRLs in this paper

\section{Experiments and Analysis}
To assess the performance of our approach, we perform experiments in meta RL scenarios and analyze its traits. 
Note that in our settings, occasions of meta learning correspond to \textit{diverse complex dynamics} and these dynamics are \textit{task conditioned}.

In all MBMRL related experiments, meta training and testing phases respectively follow that in Algorithm (\ref{Meta_Tr}) and (\ref{Meta_Te}), where \textbf{DM} is abbreviated for Dynamics Model. It is worth noting that for other MBMRL models, except GSSM/L2A\footnote{We use the same implementation in L2A \cite{nagabandi2019learning}, each learned dynamics model after fast adaptations via gradient updates is used to plan separately.}, a policy $\pi_\varphi$ is optimized across a collection of approximate dynamics models in Algorithm (\ref{Meta_Tr}) and in testing processes this policy is fine-tuned via \textit{policy gradient updates} as \textit{fast adaptations} in separate dynamics models as that in Algorithm (\ref{Meta_Te}).
Here we employ several baseline algorithms as follows:
\begin{itemize}
    \item \textbf{L2A} \cite{nagabandi2019learning}. As a gradient-based meta RL approach, the Learning to Adapt (L2A) utilizes a MAML paradigm to learn dynamics and adaptation strategies. 
    \item \textbf{MLSM-v0} \cite{galashov2019meta}. The Meta Learning Surrogate Model (MLSM) makes use of neural processes in MBMRL, where contextual latent variables are incorporated to identify different tasks.
    \item \textbf{MLSM-v1} \cite{galashov2019meta,kim2019attentive}. This is a boosted version of MLSM-v0, where an attention neural network is used in neural processes to learn sample dependent memory variables, which are aggregated into the input.
    \item \textbf{M-DPILCO} \cite{gal2016improving}. The deep PILCO, which employs Bayesian neural networks (BNNs) to capture dynamics, is meta-trained by ranging over MDPs. And the ensemble of trajectories from BNNs is used for policy optimization.
\end{itemize}

With these models, we investigate two model-based policy search strategies to combine: (i) \textit{direct policy search} trained via BPTT \cite{parmas2018pipps} (only applied to Cart-Pole environments) (ii) \textit{actor-critic policy search} using proximal policy optimization (PPO) \cite{schulman2017proximal} (applied to the rest of environments).

Meanwhile, a model-free RL algorithm is considered (referred to as \textbf{DR-PPO}), where PPO is trained across tasks as Domain Randomization. 
Another algorithm as the probabilistic embedding for actor-critic RL (referred to as \textbf{PE-PPO}) is also included in comparisons, which is a \textbf{PEARL}-like algorithm \cite{rakelly2019efficient}.
More details on environments as well settings refer to Appendix (\ref{Append_D}).

%Meta-testing algorithm in most of MBMRL
\begin{algorithm}[H]
\SetAlgoLined
\SetKwInOut{Input}{Input}
\SetKwInOut{Output}{Output}
\Input{Meta-trained $\phi_*$, $\theta_*$ and $\varphi$ ; Memory buffer $\mathcal{B}$ ;\\
Learning rate $\alpha$ ; Steps of adaptation $K$.}
\Output{Average cumulative rewards of episodes.}
Sample a testing task $\mathcal{M}_{*}{\sim}\rho(\mathcal{M})$\;

\eIf{\text{Use GSSM}}{
Run $\pi_{e}$ to collect the memory $\tau$ : $\mathcal{B}\leftarrow\mathcal{B}\cup\{\tau\}$\;

Evaluate $\pi_\varphi(a\vert s,z)$ with $z\sim q(z_c\vert \mathcal{B})$ in Eq. (\ref{mp_intra}) in $\mathcal{M}_{*}$.}
{Run $\pi_{e}$ to collect the memory $\tau$ : $\mathcal{B}\leftarrow\mathcal{B}\cup\{\tau\}$\;

\For{$i=1,2,\dots,K$}{
Sample an initial state from the learned $\text{DM}_*$\;

Collect an episode in $\text{DM}_*$ using $\pi_{\varphi}(a\vert s)$\;

Evaluate cumulative rewards in Eq. (\ref{meta_pg})\;

Optimize the policy $\pi_{\varphi}$: $\varphi\leftarrow\varphi+\alpha\nabla_{\varphi}\mathcal{J}$.}\;
Evaluate fine-tuned $\pi_\varphi(a\vert s)$ in $\mathcal{M}_{*}$.}
\caption{Meta-Testing Phases.}
\label{Meta_Te}
\end{algorithm}

\subsection{Classical Control Systems}
At first, two classical control systems are studied respectively as Cart-Pole Swing-up Systems and Acrobot Systems. And meta tasks are generated following the way in \cite{galashov2019meta,killian2017robust}. Results in Fig. (\ref{Meta_train_results}.A/B) are averaged rewards denominated by horizons.

\begin{table*}[t]
\caption{Mean Square Errors (MSEs) in Learned Dynamics Models (\texttt{DMs}) and Average Rewards in Policy Networks (\texttt{PNs}) respectively from a collection of testing environments. These results are collected using previously meta-trained models. (For each testing task, 50 episodes are sampled and averaged in rewards. Figures in brackets are standard deviations across testing tasks, and bold ones are the best.)}
\vskip 0.15in
\begin{center}
\begin{small}
\begin{sc}
\begin{tabular}{llllllll}
\toprule[1pt]
Environment & &GSSM(Ours) &M-DPILCO &MLSM-v0 &MLSM-v1 &L2A \\
\midrule[1pt]
Cart-Pole &\texttt{DM} &\textbf{0.0291}\textbf{(4.0e-2)} &0.0475(5.1e-2) &0.0626(8.1e-2) &0.0310(3.6e-2) &0.0397(4.0e-2) \\
    \cmidrule[0.55pt]{2-7}
        &\texttt{PN} &-0.5957(5.8e-2) &-0.6354(7.7e-2) &-0.6406(9.8e-2) &\textbf{-0.5935(8.2e-2)} &-0.892(3.3e-2) \\

    \midrule[1pt]
Acrobot &\texttt{DM} &\textbf{0.0022}\textbf{(8.2e-4)} &0.0056(3.9e-3) &0.0058(1.1e-3) &0.0023(1.3e-3) &0.0039(1.7e-3) \\
    \cmidrule[0.55pt]{2-7}
        &\texttt{PN} &\textbf{-0.4658}\textbf{(6.1e-2)} &-0.9444(8.9e-2) &-0.5524(9.6e-2) &-0.5286(5.3e-2) &-0.7775(5.4e-2) \\

    \midrule[1pt]
H-Cheetah &\texttt{DM} &\textbf{0.530}\textbf{(2.2E-1)} &0.678(1.4E-1) &0.533(1.4E-1) &0.636(1.4E-1) &0.785(8.4E-2) \\
    \cmidrule[0.55pt]{2-7}
        &\texttt{PN} &\textbf{1597.4}\textbf{(2.0E02)} &862.0(2.8E02) &827.3(1.9E02) &1226.8(6.4E01) &-17.9(1.3E01) \\

    \midrule[1pt]
S-Humanoid &\texttt{DM} &1.680(1.9E-1) &1.991(3.4E-1) &1.728(1.6E-1) &\textbf{1.638}\textbf{(1.9E-1)} &2.364(7.8E-2) \\
    \cmidrule[0.55pt]{2-7}
        &\texttt{PN} &\textbf{1658.6}\textbf{(1.1E02)} &1181.3(5.4E01) &485.5(8.5E01) &780.4(1.7E02) &124.9(5.7E02) \\

\bottomrule[1pt]
\end{tabular}
\end{sc}
\end{small}
\end{center}
\vskip -0.2in
\label{Meta_test_results}
\end{table*}

\textbf{Cart-Pole Results.} The physics system can be found in \cite{gal2016improving}. And we randomize masses of a cart $m_c$ and a pole $m_p$ with uniform distributions $\mathcal{U}[1.0,2.0]$ and $\mathcal{U}[0.7,1.0]$. The mission is to perform actions to reach the goal with the end of the pole. The state is $[x_c,\theta,x_c^{\prime},\theta^{\prime}]$, while the action space as the force to impose is in a continuous interval $a\sim[-10,+10]$ $N$. And the horizon in episodes is set as 25 the same as that in former works.

In Cart-Pole tasks, learning curves of GSSM and MLSM-v1 are quite similar to each other in Fig. (\ref{Meta_train_results}.A). And as listed in Table (\ref{Meta_test_results}), 50 unseen tasks are sampled to validate the performance (each task with 50 episodes). GSSM can better forecasts dynamics across tasks. In terms of policy performance, GSSM requires no additional time to adapt policies while retaining the same performance as MLSM-v1.
In addition, required samples of all MBMRL baselines in meta-training are even 2x smaller than model-free ones \cite{lillicrap2016continuous} to train in one single MDP.

\textbf{Acrobot Results.} The physics system refers to that in \cite{sutton2018reinforcement}. Similar to that in \cite{killian2017robust}, masses $m$ of two pendulums are respectively drawn from uniform distributions as $\mathcal{U}[0.8,1.2]$ and $\mathcal{U}[0.8,1.2]$. With continuous states $[\theta_1,\theta^{\prime}_1,\theta_2,\theta^{\prime}_2]$ as angles and instant angle velocities, the goal is to sequentially select an action from $\{-1, 0, +1\}$ (respectively Right Torque, No Torque, Left Torque) to reach the height above the top of the pendulum as early as possible. And the horizon is selected as 200 steps. 

In Acrobot tasks, both learning curves and testing results in Fig. (\ref{Meta_train_results}.B) demonstrate superior policy performance using GSSM. In meta testing phases, 35 unseen tasks are sampled to evaluate the model. Although in Table (\ref{Meta_test_results}) difference of dynamics approximation errors between GSSM and MLSM-v1 is tiny, the use of latent variables in GSSM advances the generalization in policies significantly. PPO-based MFMRL algorithms in Table (\ref{Meta_test_mfrl}) works a bit better than MBMRL ones but at the cost of 2x more time-steps in meta-training (600 episodes used in meta-training model-free ones). Interestingly, the use of latent variables in PE-PPO improves the performance.

Empirical observations in Fig. (\ref{Meta_train_results}.A/B) and Table (\ref{Meta_test_results}) show MLSM-v1 with lower dynamics approximation errors can attain higher rewards than other baselines. For L2A, dynamics models' approximation results are intermediate but policy performance is quite sensitive. Looking back to \textbf{Theorem 1}, we can discover that latent variable conditioned policies in GSSM lead to a \textit{tighter bound} in a meta model-based policy sense by comparing MLSM-v1 with similar dynamics predictive capability. 

\begin{table}[t]
\caption{Performance of Meta Model-free Policies in Meta-testing Processes. (Figures in brackets are standard deviations across testing tasks. Far more samples are required in meta-training than MBMRL cases.)}
\vskip 0.15in
\begin{center}
\begin{small}
\begin{sc}
\begin{tabular}{llll}
\toprule[1pt]
Environment  &DR-PPO &PE-PPO \\
\midrule[1pt]
Acrobot &-0.433(4.3E-2) &\textbf{-0.420(5.0E-2)} \\
\midrule[0.55pt]
H-Cheetah &\textbf{1360.5(1.3E02)} &608.2(7.3E01) \\
\midrule[0.55pt]
S-Humanoid &\textbf{3533.3(1.1E02)} &1248.1(1.5E02) \\
\bottomrule[1pt]
\end{tabular}
\end{sc}
\end{small}
\end{center}
\vskip -0.2in
\label{Meta_test_mfrl}
\end{table}

\subsection{Robotic Simulation Systems}
Further explorations are conducted in a robotic simulation system, which is a model-based physics engine with multi-joint dynamics known as Mujoco \cite{todorov2012mujoco}. Here we take agents of Half-Cheetah and Slim-Humanoid as instance, where the default horizon is 1000 steps. And 16 unseen tasks are sampled in meta-testing phases for each.

\begin{figure}[ht]
\centering
\includegraphics[width=0.35\textwidth]{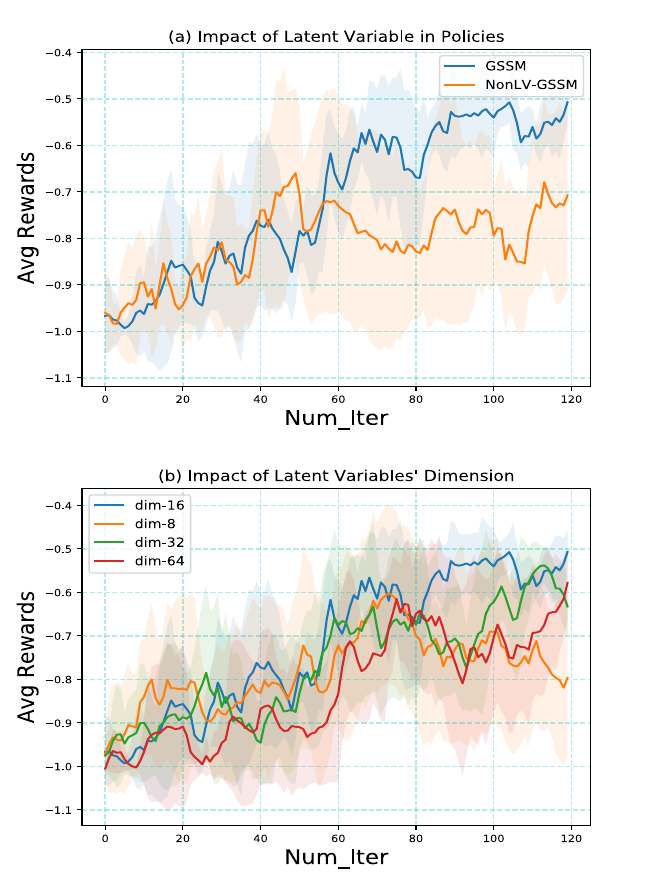}
\vspace*{-3mm}
\caption{Ablation Analysis of Policy Performance in the Acrobot's Meta-Training/Testing Process. In (a)/(b), environments are varied in terms of dynamics during iterations and each \texttt{iter} corresponds to one new sampled \texttt{episode} in updates. Shadow regions are deviations of performance in 5 runs.}
\label{Ablation_Acrobot}
\end{figure}

\textbf{Meta-HalfCheetah Results.} As exhibited in Fig. (\ref{Meta_train_results}.C), GSSM gradually improves its performance in the early stage and then surpasses all other models after 60 iterations, reaching highest average rewards at approximately 1700 level. MLSM-v0 and MLSM-v1 behaves similar trends in training, while M-DPILCO fluctuates fiercely.
In Table (\ref{Meta_test_results}) H-Cheetah, it summarizes meta-testing results across unseen tasks. GSSM is still leading in policy performance with more than 300 average rewards advantage over MLSM-v1. We also notice L2A works not so well in our environments even after trying several hyper-parameters, similar to observations in the work \cite{hiraoka2020meta,lee2020context}, and lower rewards could be due to unstable adaptations in dynamics models. 

\textbf{Meta-SlimHumanoid Results.} In Fig. (\ref{Meta_train_results}.D), learning curves reveal similar facts as that in Meta-HalfCheetah. We notice most of the models climb quickly in the early stage but these models except GSSM get stuck in a relatively sub-optimal solution, fluctuating between 500 and 700 in average rewards. And GSSM continues to rise until reaching a convergence level at 1000 average rewards. 
In Table (\ref{Meta_test_results}) S-Humanoid, both MLSM-v1 and GSSM achieve superior performance in learning dynamics but GSSM shows highest rewards and there seems no strong consistency between dynamics learning and policy learning effects in this case. 

In contrast to results in Table (\ref{Meta_test_mfrl}) H-Cheetah/S-Humanoid, where 5000 episodes are cost to train model-free ones, all MBMRL algorithms use 10x fewer samples in training processes. Especially, GSSM in Half-Cheetah environments even outperforms DR-PPO with approximately 200 in averaged rewards. Admittedly, DR-PPO in S-Humanoid achieves best performance in Table (\ref{Meta_test_mfrl}) but is less sample efficient. Another interesting discovery is PE-PPO performs worse than DR-PPO in two environments, even though probabilistic embeddings of tasks join the policy learning. Here we use permutation invariant \textit{amortized} distributions to formulate $\pi(a\vert s,z)$ as that in PEARL \cite{rakelly2019efficient}. By comparing with GSSM, where latent variables are captured from learning task dynamics, we can see the influence of representing latent variables in policy learning.  

\subsection{Ablation Studies}
To gain more insights into GSSM, two potential factors are investigated in ablation studies. 

\textbf{Role of Latent Variables.} We train GSSM using non-amortized policies as a comparison.
That is to remove the latent variable from a policy network, where a policy is parameterized as $\pi_\varphi(a\vert s)$ (Refer to Algorithm (\ref{Meta_Te}) \textit{Not Use GSSM} cases). As illustrated in Fig. (\ref{Ablation_Acrobot}.A), the evaluated non-amortized policy's performance resembles that in MLSM-v0/MLSM-v1 but significantly worse than \textit{amortized} policies $\pi_\varphi(a\vert s,z)$. The gap in Fig. (\ref{Ablation_Acrobot}.A) originates from lack of task relevant information in learning policies and the use of latent variables in a meta model-based policy sense might work as a role in effective exploration \cite{rakelly2019efficient} and system identification \cite{galashov2019meta}.

\textbf{Sensitivity Analysis.} Here dimensions of latent variables $z$ are varied in a reasonable range as $\{16,32,64\}$, and structures of GSSM are maintained in terms of dynamics models and policy networks. In Fig. (\ref{Ablation_Acrobot}.B), we notice early trends are similar before 80 iterations. With the process of optimization, the performance differences due to bottleneck dimensions gradually appear. Hence, the dimension slightly impacts results but a general trend in training stay the same. This inspires us to carefully select bottlenecks for \textit{amortized policies} in practice.

\section{Discussion and Conclusion}
In this paper, we have proposed a novel model as GSSM to combine MBRL and amortized policy learning in meta learning scenarios, where latent variables are involved in both \textit{dynamics models} and \textit{policy networks}. Our approach takes advantage of posterior sampling in decision-making. With posteriors of latent variables updated, learning performance can be gradually boosted in MBMRL.  

GSSM demonstrates the effectiveness of a graph structured model in terms of capturing task-specific system dynamics and exhibits superior generalization across tasks. In particular, \textit{amortized policies} learned in GSSM allow for \textit{fast adaptations} to new tasks without \textit{additional policy gradient updates}. This kind of trait helps us avoid either re-planning or adaptations in policies, making our approach fascinating in missions when planning time is quite sensitive.

%\newpage
%\section*{Acknowledgements}
%Q.W. gratefully acknowledges sincere and consistent support from Prof. Max Welling during his pursuing Ph.D at AMLAB. Also, we thank NVIDIA Corporation for the donation of a Titan V GPU.
\bibliography{example_paper}
\bibliographystyle{icml2020}

\appendix
\newpage
\onecolumn
\icmltitle{Supplementary Materials}

\renewcommand\thesection{\Alph{section}}

\section{Proof of Lemma 1}\label{Append_A}
This proof is based on partial results in \cite{rajeswaran2020game}, and an extension is performed here. The context is set in a distribution over MDPs $M\sim p(M)$, and a sampled real environment and the corresponding learned dynamics model are respectively denoted as $M$ and $\hat{M}$.

\textbf{Corollary 1.} Assuming a single step reward in a Markov Decision Process has a supremum value $\mathcal{R}_{\max}$ and the discounted factor for rewards $\gamma<1$, then the state value function $V^{\pi}(s)$ under a policy $\pi$ can be bounded with the following inequality.
\begin{equation}
    \begin{split}
        \max_{s\in\mathcal{S}} V^{\pi}(s)\leq\frac{\mathcal{R}_{\max}}{1-\gamma},\quad \forall\pi
    \end{split}
\end{equation}
\textbf{Proof.} The state value $V^{\pi}(s)$ can be computed in the form of $\mathbb{E}_{\pi}\big[\sum_{t=0}^{\infty}\gamma^{t}\mathcal{R}_t\vert S_0=s\big]=\int p(\tau)\mathcal{R}(\tau)d{\tau}$, where the cumulative reward for trajectory $\tau$ is $\mathcal{R}(\tau)=\sum_{t=0}^{\infty}\gamma^{t}\mathcal{R}_t$ with an initial state $S_0=s$. Also note that $\sup \{\mathcal{R}_t\}\leq\mathcal{R}_{\max}$, it is trivial to verify the equation.
\begin{equation}
    \begin{split}
    \mathcal{R}(\tau)\leq(\sum_{t=0}^{\infty}\gamma^{t})\mathcal{R}_{\max}=\frac{\mathcal{R}_{\max}}{1-\gamma} \\
    V^{\pi}(s)=\int p(\tau)\mathcal{R}(\tau)d{\tau}\leq\frac{\mathcal{R}_{\max}}{1-\gamma}
    \end{split}
\end{equation}

Note the Bellman equation in terms of any state value $V_{\pi}^{M}(s)$ under a policy $\pi$ in a dynamics model $M$,
\begin{equation}
    \begin{split}
        V_{\pi}^{M}(s)=\int \big(r(s,a,s^{\prime})+\gamma V_{\pi}^{M}(s^{\prime})\big)\pi(a\vert s)p(s^{\prime}\vert s,a)dads^{\prime}
        =\mathcal{R}_{\pi}^{M}(s)+\gamma\mathbb{E}_{s^{\prime}\sim p_{\pi}^{M}(\cdot\vert s)}\big[V_{\pi}^{M}(s^{\prime})\big]
    \end{split}
\end{equation}
where $\mathcal{R}_{\pi}^{M}(s)$ is the expected one step rewards and $p_{\pi}^{M}(\cdot\vert s)$ is the state transition distribution and both depend on the environment and the policy.

Hence, we can naturally estimate the difference between state values in two mentioned dynamics models with the help of \textbf{Corollary 1} as follows.
\begin{equation}
    \begin{split}
        |V_{\pi}^{M}(s)-V_{\pi}^{\hat{M}}(s)|\leq
        |\mathcal{R}_{\pi}^{M}(s)-\mathcal{R}_{\pi}^{\hat{M}}(s)|+\gamma|\mathbb{E}_{s^{\prime}\sim p_{\pi}^{M}(\cdot\vert s)}\big[V_{\pi}^{M}(s^{\prime})\big]-\mathbb{E}_{s^{\prime}\sim p_{\pi}^{\hat{M}}(\cdot\vert s)}\big[V_{\pi}^{\hat{M}}(s^{\prime})\big]| \\
        \leq 2\mathcal{R}_{\max}D_{\text{TV}}[p_{\pi}^{\hat{M}}(\cdot\vert s),p_{\pi}^{M}(\cdot\vert s)]+\gamma|\mathbb{E}_{s^{\prime}\sim p_{\pi}^{M}(\cdot\vert s)}\big[V_{\pi}^{M}(s^{\prime})\big]-\mathbb{E}_{s^{\prime}\sim p_{\pi}^{\hat{M}}(\cdot\vert s)}\big[V_{\pi}^{M}(s^{\prime})\big]| \\
        +\gamma|\mathbb{E}_{s^{\prime}\sim p_{\pi}^{\hat{M}}(\cdot\vert s)}\big[V_{\pi}^{M}(s^{\prime})\big]-\mathbb{E}_{s^{\prime}\sim p_{\pi}^{\hat{M}}(\cdot\vert s)}\big[V_{\pi}^{\hat{M}}(s^{\prime})\big]| \\
        \leq 2\mathcal{R}_{\max}D_{\text{TV}}[p_{\pi}^{\hat{M}}(\cdot\vert s),p_{\pi}^{M}(\cdot\vert s)]+2\gamma\big(\max_{s^{\prime}}V_{\pi}^{M}(s^{\prime})\big)D_{\text{TV}}[p_{\pi}^{\hat{M}}(\cdot\vert s),p_{\pi}^{M}(\cdot\vert s)] \\
        +\gamma\max_{s^{\prime}}|V_{\pi}^{M}(s^{\prime})-V_{\pi}^{\hat{M}}(s^{\prime})|,\quad \forall s\in\mathcal{S}
    \end{split}
\end{equation}

Since the left side term is satisfied for all states, we can naturally have the following equation.
\begin{equation}
    \begin{split}
        (1-\gamma)\max_{s^{\prime}}|V_{\pi}^{M}(s^{\prime})-V_{\pi}^{\hat{M}}(s^{\prime})|\leq
        2\big(\mathcal{R}_{\max}+\gamma\big(\max_{s^{\prime}}V_{\pi}^{M}(s^{\prime})\big)\big)D_{\text{TV}}[p_{\pi}^{\hat{M}}(\cdot\vert s),p_{\pi}^{M}(\cdot\vert s)]
    \end{split}
    \label{sing_gap}
\end{equation}
Then by imposing $\mathbb{E}_{M\sim p(M)}$ over both sides in Eq. (\ref{sing_gap}) and with the meta dynamics model approximated error $\mathbb{E}_{M\sim p(M)\atop (s,a)\sim\nu(s,a)}\big[D_{\text{TV}}[\hat{P}_{M}(\cdot\vert s,a),P_{M}(\cdot\vert s,a)]\big]\leq\epsilon$, we can give the regret bound as follows.
\begin{equation}
    \begin{split}
    \mathbb{E}_{M\sim p(M)}\big[\max_{s^{\prime}}|V_{\pi}^{M}(s^{\prime})-V_{\pi}^{\hat{M}}(s^{\prime})|\big]\leq
    \frac{2\epsilon\mathcal{R}_{\max}}{(1-\gamma)^2},\quad
    \forall\pi
    \end{split}
\end{equation}
Finally, the performance gap can be measured with Eq. (\ref{fina_gap}), and \textbf{Lemma 1} is proved.
\begin{equation}
    \begin{split}
        \mathbb{E}_{M\sim p(M)}\big[|\mathcal{J}_{\hat{M}}(\pi)-\mathcal{J}_M(\pi)|\big]\leq
    \frac{2\epsilon\mathcal{R}_{\max}}{(1-\gamma)^2},\quad
    \forall\pi
    \end{split}
    \label{fina_gap}
\end{equation}

\section{Proof of Theorem 1}\label{Append_B}
Here let us refer to optimal policies in an arbitrary MDP $\mathcal{M}$ and its approximation $\mathcal{\hat{M}}$ as $\pi_M$ and $\pi_{\hat{M}}$ respectively. With the induction in \textbf{Lemma 1}, we reuse Eq. (\ref{sing_gap}) and it is trivial to verify the following equations.

\begin{equation}
    \begin{split}
        |\mathcal{J}_{\hat{M}}(\pi)-\mathcal{J}_M(\pi)|\leq
        \frac{2\mathcal{R}_{\max}D_{\text{TV}}[p_{\pi}^{\hat{M}}(\cdot\vert s),p_{\pi}^{M}(\cdot\vert s)]}{(1-\gamma)^2},\quad
        \forall\pi
    \end{split}
\end{equation}

Hence, we can have the inequality based on the truth that $\pi_{\hat{M}}$ is optimal in $\mathcal{\hat{M}}$ and reuse Eq. (\ref{sing_gap}).

\begin{equation}
    \begin{split}
        \mathcal{J}_M(\pi_M)\leq 
        \mathcal{J}_{\hat{M}}(\pi_M)+\frac{2\mathcal{R}_{\max}D_{\text{TV}}[p_{\pi}^{\hat{M}}(\cdot\vert s),p_{\pi}^{M}(\cdot\vert s)]}{(1-\gamma)^2}\leq
        \mathcal{J}_{\hat{M}}(\pi_{\hat{M}})+\frac{2\mathcal{R}_{\max}D_{\text{TV}}[p_{\pi}^{\hat{M}}(\cdot\vert s),p_{\pi}^{M}(\cdot\vert s)]}{(1-\gamma)^2} \\
        \leq\mathcal{J}_M(\pi_{\hat{M}})+\frac{4\mathcal{R}_{\max}D_{\text{TV}}[p_{\pi}^{\hat{M}}(\cdot\vert s),p_{\pi}^{M}(\cdot\vert s)]}{(1-\gamma)^2}
    \end{split}
\end{equation}

With the help of expectation over the distribution of MDPs, the final equation of a lower bound can be drawn as that in \textbf{Theorem 1}.
\begin{equation}
    \begin{split}
        \mathbb{E}_{M\sim p(M)}\big[\mathcal{J}_{M}(\pi_{\hat{M}})\big]\geq
        \mathbb{E}_{M\sim p(M)}\big[\mathcal{J}_{M}(\pi_{M})\big]-\frac{4\epsilon\mathcal{R}_{\max}}{(1-\gamma)^2}
    \end{split}
\end{equation}

\section{Evidence Lower Bound for GSSM}
Here a distribution $P(D)$ describes the state action pair in meta-training processes, and each data point is attached with a memory set $[x_c,y_c]$ to imply the statistics information from a task.
With Jessen's inequality and the approximate posterior $q_{\phi}(z_*\vert x_*,x_c,y_c)$, we can have evidence lower bound as follows.
\begin{equation}
\begin{split}
    \mathbb{E}_{p(D)}\ln p(y_*\vert x_*,x_c,y_c)=\mathbb{E}_{p(D)}\ln\mathbb{E}_{q_{\phi}}\big[\frac{p(z_{*})}{q_{\phi}(z_*\vert x_*,x_c,y_c)} p_{\theta}(y_*\vert x_*,z_*)\big] \\
    \geq
    \mathbb{E}_{p(D)}\mathbb{E}_{q_\phi}\ln\big[p_{\theta}(y_*\vert x_*,z_*)\big]-\mathbb{E}_{p(D)}\mathbb{E}_{q_\phi}\ln\big[\frac{q_{\phi}(z_*\vert x_*,x_c,y_c)}{p(z_{*})}\big] 
\end{split}
\end{equation}
By replacing the zero information prior distribution with a parameterized approximate prior $q(z_*\vert x_c,y_c)$, we can derive the formerly mentioned ELBO.
\begin{equation}
    \begin{split}
        \mathbb{E}_{p(D)}\big[\ln \underbrace{p(y_*\vert x_*,x_c,y_c)}_{\text{intractable data likelihood}}\big]\geq\mathbb{E}_{p(D)}\big[\mathbb{E}_{q_{\phi_1}}[\ln p_{\theta}(y_*\vert x_*,z_*)] \\
        -D_{KL}[\underbrace{q_{\phi_1}(z_*\vert x_*,x_c,y_c)}_{\text{approximate posterior}}\parallel \underbrace{q_{\phi_2}(z_c\vert x_c,y_c)}_{\text{approximate prior}}]\big]
    \end{split}
    \label{dm_elbo_append}
\end{equation}

Note that both the approximate prior and the posterior are learnable with a partially shared neural network in meta learning scenarios, which shares similar motivations in works \cite{denton2018stochastic,pertsch2020accelerating,garnelo2018neural}. For more information on encoding relationship between these context points and target points, refer to Fig. (\ref{GSSM_Structure}).

\section{Experimental Details and Neural Architectures}\label{Append_D}
In this section, we present environments, training details, neural architectures as well as parameter settings. The anonymous download link for our codes is attached here in the peer-reviewed version.

%Especially, the code is attached here in an anonymous way (https://github.com/codeanonymous233/anonymouscode), where you can have more information about GSSM.

\subsection{Environmental Details}
\textbf{MBMRL Tasks.} Here we describe mentioned meta reinforcement learning tasks in this paper. The Cart-Pole environment can be found in the link\footnote{https://github.com/BrunoKM/deep-pilco-torch} here. The Acrobot is based on open-ai gym\footnote{https://gym.openai.com/} and Half-Cheetah/Slim-Humanoid are from a Mujoco package\footnote{http://www.mujoco.org/}. Generations of diverse Cart-Pole/Acrobot environments have been introduced in the main paper.
As for configurations of Half-Cheetah/Slim-Humanoid environments, we generate the Meta-training MDPs via the combination of the mass re-scaled coefficient in the list $\{0.8, 0.9, 1.0, 1.1, 1.2\}$ and the damping coefficient in the list $\{0.8, 0.9, 1.0, 1.1, 1.2\}$, while those hyper-parameters for Meta-testing phases are $\{0.85, 0.95, 1.05, 1.15\}$ for both mass-rescaled and damping coefficients. As a result, totally 16 unseen MDPs are generated by the Cartesian of mass coefficients and damping coefficients for meta-testing processes.

\begin{figure}[ht]
\centering
\includegraphics[width=0.85\textwidth]{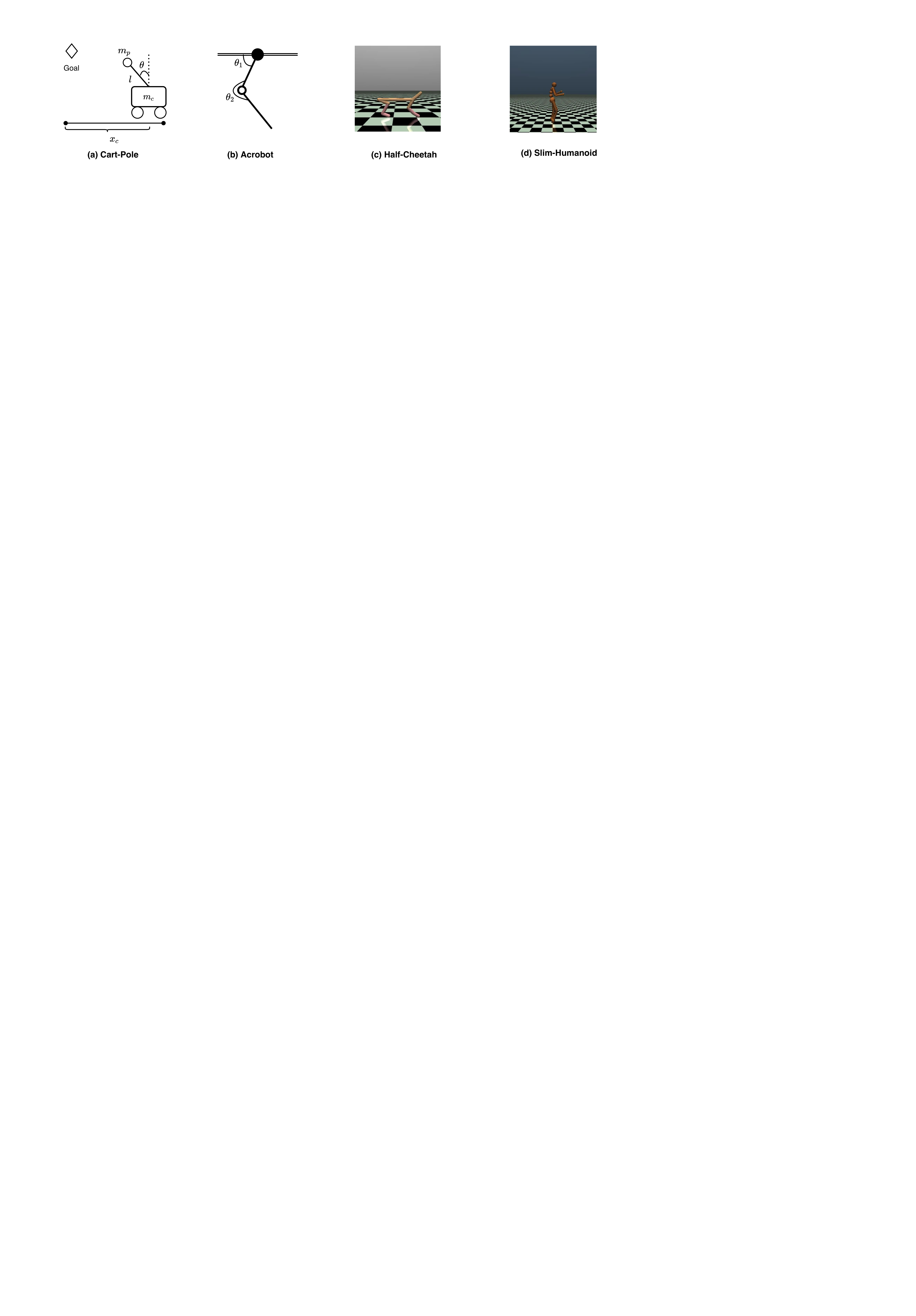}
\caption{Fundamental Environments used in Meta Model-based Reinforcement Learning Experiments.}
\label{Meta_RL_Envs}
\end{figure}

\textbf{Reward Descriptions.} Besides, reward functions are listed here (Refer to Table (\ref{env_reward_function})). More details are as follows. In Cart-Pole environments, $d$ in a reward function measures the square of the distance between the pole's end point and its goal, and hyper-parameter $\sigma_c=0.25$. In Acrobot environments, the list of parameters $\{l_1,l_2,\theta_1,\theta_2\}$ refers to Fig. (\ref{Meta_RL_Envs}) in terms of meanings in a reward function. In Half-Cheetah environments, $x_t$ is the notation of the x-coordinate in the Half-Cheetah agent at time slot index $t$, $\nabla_t$ is time difference in dynamics (the resulted ratio is the speed of agent.) and $a_t$ is the action performed instantly. In Slim-Humanoid environments, notations are similar to those in Half-Cheetah and $x_{t,h}$ in reward functions refer to the instant torso's height. Horizons of trajectories as well as types of action spaces can also be found in Table (\ref{env_reward_function}). 

\begin{table*}[t]
\caption{Reward Functions in Related Environments. }
\label{env_reward_function}
\vskip 0.15in
\begin{center}
\begin{small}
\begin{sc}
\begin{tabular}{llll}
\toprule
Environments &Reward Functions &Horizon &Control\\
\midrule
Cart-Pole &$1-\exp{(-\frac{\|d^2\|}{\sigma_{c}^{2}})}$ & 25 &Continuous\\
\midrule
Acrobot &bool($-l_{1}\cos{(\theta_1)}-l_{2}\cos{(\theta_{1}+\theta_{2})-l_{1}}$) & 200 &Discrete\\
\midrule
Half-Cheetah &$\frac{x_{t+1}-x_{t}}{\nabla t}-0.1*\|a_t\|_{2}^{2}$ & 1000 &Continuous\\
\midrule
Slim-Humanoid &$\frac{50(x_{t+1}-x_{t})}{3\nabla t}-0.1*\|a_t\|_{2}^{2}+5.0*\text{bool}(1.0\leq x_{t,h}\leq 2.0)$ & 1000 &Continuous\\
\bottomrule
\end{tabular}
\end{sc}
\end{small}
\end{center}
\vskip -0.1in
\end{table*}

\begin{figure*}[ht]
\centering
\includegraphics[width=0.9\textwidth]{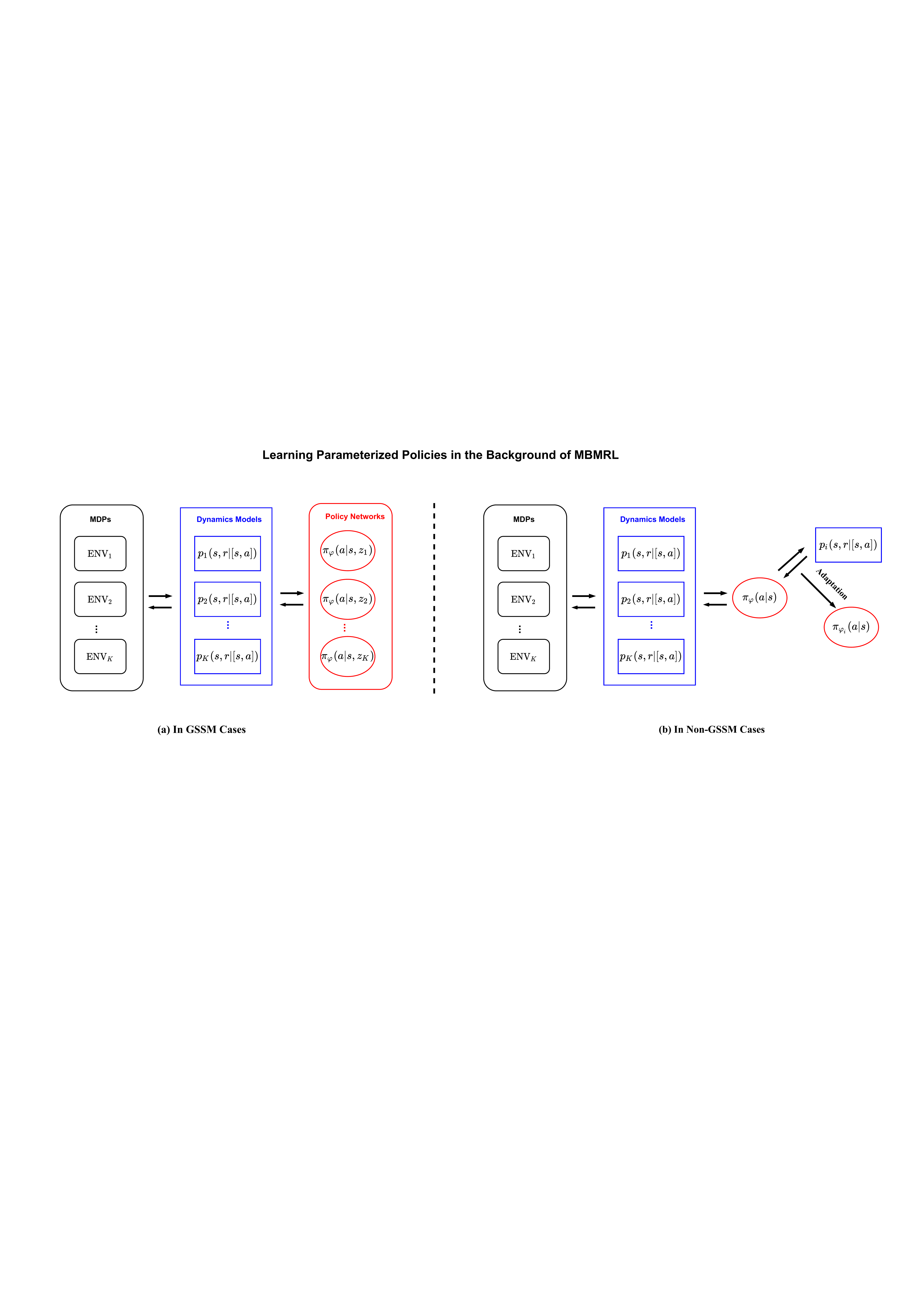}
%\vspace*{-5mm}
\caption{Meta Model-based Policy Search used in Models. Note that reducing adaptation time in policies is the first priority in policy search strategies in this work. In the \textbf{Left}, amortized policies are used and latent variables are to specify different tasks. In the \textbf{Right}, the meta-trained policy is not conditioned on latent variables and needs to be adapted to respective tasks.}
\label{MBMRL_PS}
\end{figure*}

\subsection{Training Details}
\textbf{Data Preprocessing.} In Acrobot tasks, the output of dynamics models is the next or transited state ($x=[s,a]$, $y=s^{\prime}$ ). In other tasks, the output of dynamics models is the difference of the next state and the current state ($x=[s,a]$, $y=\Delta s$). For the input of dynamics models, it is the state-action pair in all environments. For Half-Cheetah/Slim-Humanoid environments, standardization is required for both input and output of dynamics models during training processes.
 
\textbf{More Details in Policy Search.} In Cart-Pole Swing-Up environments, Back-Propagation Through Time (BPTT) is used in model-based policy search and the policy network as a radial basis function follows that in \cite{gal2016improving}, except for GSSM a latent variable is concatenated in the input. In Acrobot/Half-Cheetah/Slim-Humanoid, we combine PPO with the learned dynamics model (We also perform additional trials in BPTT strategies but \textit{this kind of model-based policy search suffers from gradient exploding in practice}), and a direct combination of model-based and model-free RL algorithms in meta-learning leads to stable training.

Besides, meta-trained policies in MLSM-v0/MLSM-v1/M-DPILCO require additional policy gradient updates in separate dynamics models of tasks and these are up to tasks based on our trials: for Cart-Pole/Acrobot, five trajectories are enough to fine-tune these policies, and for Half-Cheetah/Slim-Humanoid one trajectory is enough to fine-tune these policies. Due to nature of model-based policy search, additional training might result in over-fitting in a policy level. Here we report required adaptation time of policies across tasks using different models during meta-testing processes in Table (\ref{Meta_Adapt_Time}) and the unit is in seconds.

All of these are reflected in Fig. (\ref{MBMRL_PS}). 
Traditional model predictive control strategies are prohibitively expensive in implementations, \textit{costing much more time with lower efficiency in high dimensional action space}. Our implementation is a first-step trial but well matches practical demands in the future.
\textit{Since related work employing parameterized policies in MBMRL remains limited, our implementation in this domain is a preliminary exploration.}

\textbf{Further Descriptions in Fig.s/Tables.}
Here we need to add more descriptions on Cart-Pole, where authors can follow the implementations in the work\footnote{https://github.com/BrunoKM/deep-pilco-torch}, and the state-of-art performance using Deep-PILCO is about -0.6 in episodes for a single task. We also try DR-PPO and PE-PPO in Cart-Pole tasks with more than 10x required time steps in training, but the resulted performance in testing is far worse than MBMRL ones and we guess the PPO algorithm here cannot well handle planning with short horizons (Other referred model-free results can be found in \cite{lillicrap2016continuous}).

The Fig. (\ref{Meta_train_results}) keeps track of meta-training performance using MBMRL algorithms, and dynamics of MDPs are changed with iterations. And every fixed number of iterations, MDPs are resampled with time (for Cart-Pole, every 10 iters; for Acrobot, every 3 iters; for mujoco ones, every 2 iters).
The Table (\ref{Meta_test_results}) summarizes the meta-testing results over unseen MDPs. Some additional explanations are as follows.
In meta-testing tasks of Cart-Pole, contextual latent variables in GSSM/MLSM-v0/MLSM-v1 are computed after transitions of two trajectories (50 transition steps) are aggregated. 
In meta-testing tasks of Acrobot, contextual latent variables in GSSM/MLSM-v0/MLSM-v1 are computed after transitions of a one-sixth trajectory (50 transition steps) are aggregated. 
In meta-testing tasks of Half-Cheetah/Slim-Humanoid (Refer to results in Table (\ref{Meta_test_results})), contextual latent variables in GSSM/MLSM-v0/MLSM-v1 are computed after transitions of a half trajectory (500 transition steps) are aggregated.

\begin{figure*}[ht]
\centering
\includegraphics[width=0.8\textwidth]{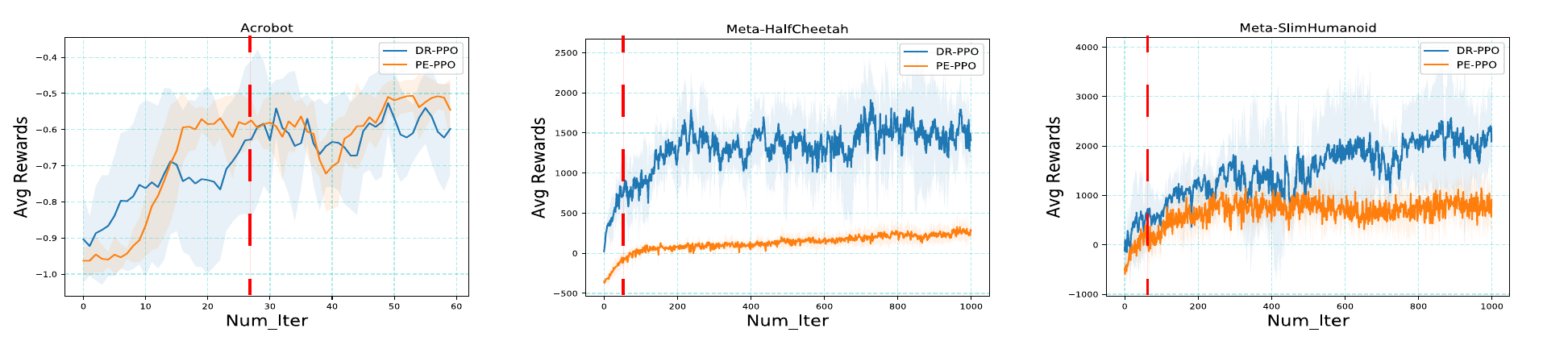}
%\vspace*{-5mm}
\caption{Performance of Policies in Meta-training Model-Free Reinforcement Learning. From left to right are respectively Acrobot, HalfCheetah and SlimHumanoid, and every \texttt{Iter} correspond to 10 episodes for Acrobot, 5 episodes for HalfCheetah/SlimHumanoid. Red vertical dotted lines measure the threshold of required time-steps used to train MBMRL baselines in our paper. 5 runs are performed to averaged in experiments.}
\label{meta_mfrl}
\end{figure*}

Meanwhile meta-training processes in model-free meta reinforcement learning are recorded in Fig. (\ref{meta_mfrl}) and note that with the same quantity of samples as that used in MBMRL ones (red vertical dotted lines are measures of required samples in MBMRL), results in curves are far  worse than those in MBMRL ones. These are trained with Adam optimizers and learning rates are 5e-4 in default.

\subsection{Neural Architectures and Parameter Settings}
Here neural architectures in meta dynamics models are listed in Table (\ref{meta_dm_nn}). These architectures are shared across all implemented tasks in the paper. And one layer graph encoding is enough to guarantee performance in our GSSM implementations for all experiments. For Meta-DPILCO, neural architectures resemble that in the table except that encoders for latent variables are removed and dropout modules are integrated in each layer. In Cart-Pole environments, parameters in Table (\ref{meta_dm_nn}) are $\{n=2, dim\_latxy=32, dim\_lat=16, m=2, dim\_h=200\}$. In Acrobot environments, parameters in Table (\ref{meta_dm_nn}) are $\{n=2, dim\_latxy=32, dim\_lat=16, m=5, dim\_h=400\}$. In Mujoco environments, parameters in Table (\ref{meta_dm_nn}) are $\{n=2, dim\_latxy=32, dim\_lat=16, m=5, dim\_h=400\}$. Interestingly, our GSSM requires less parameters than those in MLSM-v1 because of encoder structures (GSSM and MLSM-v0 share similar scales of model parameters).

As for meta policy networks or latent variable conditioned policy networks (used in GSSM), we adopt the ordinary ones and these are listed in Table (\ref{meta_pn}). In Cart-Pole environments, parameters in Table (\ref{meta_pn}) are $\{n_p=1, dim\_ph=50\}$. In Acrobot environments, parameters in Table (\ref{meta_pn}) are $\{n_{pa}=1, dim\_ph=128, n_{pc}=1\}$.
In Half-Cheetah environments, parameters in Table (\ref{meta_pn}) are $\{n_{pa}=1, dim\_ph=128, n_{pc}=1\}$. In model-free meta reinforcement learning scenarios, the contextual encoder is permutation invariant the same as that used in MLSM-v0, and the optimization objectives follow those in PEARL \cite{rakelly2019efficient}.

\begin{table*}[t]
\caption{Neural Network Structure of MBMRL Models. The transformations in the table are linear, followed with ReLU activation mostly. As for MLSM-v1, the encoder network is doubled in the table since there exists a local variable for prediction. }
\label{meta_dm_nn}
\vskip 0.15in
\begin{center}
\begin{small}
\begin{sc}
\begin{tabular}{lll}
\toprule
NP Models &Encoder &Decoder \\
\midrule
        &$[dim\_x,dim\_y]\mapsto \underbrace{dim\_latxy\mapsto dim\_latxy}_{n\;times}$ &$[dim\_x,(2*)dim\_lat]\mapsto \underbrace{dim\_h\mapsto dim\_h}_{m\;times}$ \\
MLSM-v0/v1 &$dim\_latxy\mapsto dim\_lat$. &$dim\_h\mapsto dim\_y$ \\
\midrule
        &$[dim\_x,dim\_y]\mapsto \underbrace{dim\_latxy\mapsto dim\_latxy}_{n\;times}$ &$[dim\_x,dim\_lat]\mapsto \underbrace{dim\_h\mapsto dim\_h}_{m\;times}$ \\
GSSM    &$dim\_x\mapsto dim\_latx$; & \\
        &$[dim\_latx,dim\_laty]\mapsto dim\_lat$. &$dim\_h\mapsto dim\_y$\\
\bottomrule
\end{tabular}
\end{sc}
\end{small}
\end{center}
\vskip -0.1in
\end{table*}

\begin{table*}[t]
\caption{Neural Network Structure in Meta Policy Networks. For Back-propagation Through Time (BPTT) and Actor-Critic Policy Gradient Algorithms, neural architectures are different. ReLU is used as an activation function. Soft-max is used in the output of Actor Network in the discrete control.}
\label{meta_pn}
\vskip 0.15in
\begin{center}
\begin{small}
\begin{sc}
\begin{tabular}{lll}
\toprule
Policy Training &Neural Architectures \\
\midrule
        &$[dim\_obs]/[dim\_obs, dim\_lat]\mapsto \underbrace{dim\_ph\mapsto dim\_ph}_{n_p\;times}$ \\
BPTT &$dim\_ph\mapsto dim\_act$. \\
\midrule
               &$[dim\_obs]/[dim\_obs, dim\_lat]\mapsto \underbrace{dim\_ph\mapsto dim\_ph}_{n_{pc}\;times}\mapsto dim\_act$ (Actor Network) \\
AC-PG (PPO) &$[dim\_obs]/[dim\_obs, dim\_lat]\mapsto \underbrace{dim\_ph\mapsto      dim\_ph}_{n_{pa}\;times}\mapsto 1$ (Critic Network). \\
\bottomrule
\end{tabular}
\end{sc}
\end{small}
\end{center}
\vskip -0.1in
\end{table*}

%For L2A, we set the batch number of task as 5 and 2000 iterations are performed, costing 10000 trajectories in total to train the model (inner and outer loop learning rates are both 1e-3).

\begin{table*}[t]
\caption{Additional Adaptation Time in Policies. Time units are in \texttt{s} (seconds). In L2A, each dynamics model after gradient adaptations is used to plan separately and this requires far more time than other MBMRL algorithms. In GSSM, no additional adaptations in policies are needed. (Figures in brackets are variances across testing tasks.)}
\vskip 0.15in
\begin{center}
\begin{small}
\begin{sc}
\begin{tabular}{lllllll}
\toprule[1pt]
Environment &GSSM(Ours) &M-DPILCO &MLSM-v0 &MLSM-v1 &L2A \\
\midrule[1pt]
Cart-Pole &$\O$ &0.26(5.3E-4) &0.34(6.1E-4) &0.55(2.5E-4) &$\ast\ast\ast$ \\
        
    \midrule
Acrobot &$\O$ &4.3(1.0E-2) &5.2(8.9E-2) &6.5(4.0E-2) &$\ast\ast\ast$ \\

    \midrule
H-Cheetah &$\O$ &2.5(9.6E-3) &2.9(1.4E-2) &4.6(2.4E-2) &$\ast\ast\ast$ \\

    \midrule
S-Humanoid &$\O$ &2.4(5.3E-3) &3.1(7.6E-3) &4.9(3.4E-2) &$\ast\ast\ast$ \\

\bottomrule[1pt]
\end{tabular}
\end{sc}
\end{small}
\end{center}
\vskip -0.2in
\label{Meta_Adapt_Time}
\end{table*}

\subsection{Computing Devices and Required Platforms}
Throughout the work, we run experiments in a GTX 1080-Ti GPU, and Pytorch\footnote{https://pytorch.org/} is used in implementations.

\end{document}